\pdfoutput=1

\documentclass[11pt]{article}

\usepackage{acl}

\usepackage{times}
\usepackage{latexsym}
\usepackage{graphicx}
\usepackage{multirow}
\usepackage{float}

\usepackage[T1]{fontenc}

\usepackage[utf8]{inputenc}

\usepackage{microtype}

\title{Generating Query-Focused Summarization Datasets from Query-Free Summarization Datasets}

\author{Yllias Chali \\
  University of Lethbridge \\
  Alberta, Canada \\
  \texttt{yllias.chali@uleth.ca} \\\And
  Deen Abdullah \\
  University of Lethbridge \\
  Alberta, Canada \\
  \texttt{deen.abdullah@uleth.ca} \\
  \texttt{} \\}

\begin{document}
\maketitle
\begin{abstract}
Large-scale datasets are widely used to perform summarization tasks, but they may not include queries alongside documents and summaries. In the search for suitable datasets for Query-Focused Summarization (QFS), we identify two research questions: Is it possible to automatically generate evidence-based query keywords from query-free datasets? Does evidence-based query generation support the QFS task? This paper proposes an evidence-based model to generate queries from query-free datasets. To evaluate our model intrinsically, we compare the similarity between the original queries and the system-generated queries of two QFS datasets. We also perform summarization tasks using different pre-trained models, as well as a state-of-the-art (SOTA) QFS model, to measure the extrinsic performance of our query generation approach. Experimental results indicate that summaries generated using evidence-based queries achieve competitive ROUGE scores compared to those generated from the original queries.
\end{abstract}

\section{Introduction}
Query-focused summarization (QFS) focuses on generating summaries from original documents, where the summary is tailored to a specific given query. Researchers have implemented various neural models and proposed distinctive approaches that advance both extractive and abstractive query-focused summarization tasks \citep{lin-2004-rouge, gupta-etal-2007-measuring, 10.5555/1625275.1625743, OUYANG2011227, 10.1145/3077136.3080690, nema-etal-2017-diversity, hasselqvist2017querybased, baumel2018query, abdullah-chali-2020-towards, xu-lapata-2020-coarse, 10.1007/978-3-030-47358-7_35, su2021improve}. However, the lack of appropriate datasets for QFS has always been a concern for researchers, as the unavailability of large-scale datasets makes the task more challenging \citep{fisher2006query, see2017point, liu-lapata-2019-text, abdullah-chali-2020-towards}. Thus, the shortage of large-scale QFS datasets motivates the development of effective query generation approaches.
To address this issue, we propose a context-oriented, evidence-based model that supports query-focused summarization by generating queries from documents in any query-free dataset. Using a transfer learning approach, we train our evidence-based model on the {article, highlight} pairs from the CNN/DailyMail dataset. To avoid data bias, we use different datasets, such as Debatepedia and TD-QFS, for the QFS task instead of CNN/DailyMail. Additionally, both datasets contain queries, allowing us to compare the performance of original queries with that of the generated evidence-based queries. Samples of the original and evidence-based queries from the TD-QFS dataset are shown in Table~1.

\begin{table}[h]
\centering
\begin{tabular}{|c|c|} \hline 
\multicolumn{2}{|c|}{Sample Queries} \\
Original & Evidence-based \\ \hline
asthma causes & asthma chronic disease \\
                         & affects air \\ \hline
Lung cancer &  lung cancer screening \\
diagnosis      & currently routine practice \\ \hline
\end{tabular}
\caption{Sample queries (original and the evidence-based) from TD-QFS dataset.}
\end{table}

\section{Related Work}
Pre-trained models, including BERT \citep{devlin-etal-2019-bert}, GPT \citep{radford2019language}, RoBERTa \citep{liu2019roberta}, T5 \citep{raffel2020exploring}, LED \citep{beltagy2020longformer}, BART \citep{lewis-etal-2020-bart}, and PEGASUS \citep{pmlr-v119-zhang20ae}, have been widely used on various datasets such as Gigaword \citep{ma-huang-2006-uniform}, CNN/DailyMail \citep{hermann2015teaching}, SQuAD \citep{rajpurkar-etal-2016-squad}, TD-QFS \citep{Baumel_Cohen_Elhadad_2016}, and Debatepedia \citep{nema-etal-2017-diversity} to perform summarization, machine translation, and other NLP tasks \citep{rush-etal-2015-neural, nallapati-etal-2016-abstractive, durrett-etal-2016-learning}.

By emphasizing a query-based attention mechanism, \citet{nema-etal-2017-diversity} implemented a diversity-driven model that reduces repetitive phrases in summaries. \citet{abdullah-chali-2020-towards} proposed a query generation approach that considers both the input document and the target summary. Similarly, \citet{xu-lapata-2020-coarse} addressed the problem of query–cluster interaction and proposed a coarse-to-fine model for query-focused multi-document extractive summarization.

In this paper, we propose an evidence-based model that leverages a transfer learning approach to generate evidence-based queries. First, we train a model to generate evidence keywords from {article, highlight} pairs in the CNN/DailyMail dataset. We then use this evidence model to generate evidence-based queries for the Debatepedia and TD-QFS datasets.

\section{Problem Definition}
Given a query $Q_{i}$ and a document $D_{i}$, we generate a query-relevant summary $S_{i}$ in the QFS task. A query should focus on the parts of the document that are related to its keywords, while the summary should cover the corresponding query-relevant contexts within the document. We define the common context words present in both the document and the summary as evidence in this work. However, the challenge lies in generating such evidence using only the document. Therefore, we hypothesize that a transfer learning approach can help train an evidence-based model for the document-to-evidence-based query generation task.

\section{Our Framework}
In query-focused summarization, the summary should align with the query, meaning that query-related information must be present in the document. Therefore, the query should be supported by both the summary and the document, implying that evidence keywords must be reflected in the query. However, only a few QFS datasets provide query–document–summary triads, and these are often constructed based on simplifying assumptions. For example, in the Debatepedia dataset, questions from controversial debates are treated as queries, while topic titles are treated as summaries. In such cases, some titles (summaries) may not be fully relevant to the queries.

Motivated by this limitation, we investigate whether evidence keywords can perform better than the original queries in QFS datasets. If our hypothesis holds, evidence-based queries could be applied to query-free resources for QFS tasks. Therefore, we propose an evidence-based model that generates evidence from documents to be used as queries. Our work\footnote{Our code is available at https://anonymous.4open.science/r/Our-Project} consists of two main steps. First, we fine-tune a pre-trained model on the CNN/DailyMail dataset for the document-to-query generation task. Then, using the evidence-based model, we generate evidence-based queries for the Debatepedia and TD-QFS datasets without accessing the summaries. This transfer learning approach helps us avoid target leakage while generating evidence-based queries.

We use the CNN/DailyMail dataset, which contains news article–highlight pairs, to generate evidence-based queries. Specifically, we extract the common words from news articles and their corresponding highlights and consider them as evidence using Equation 1:

\begin{equation} E_{i} \leftarrow \{w_{ij}\} \; (if \; w_{ij}= w_{ik}) \end{equation}

where $E_{i}$ is the set of extracted evidence keywords from the news article $(N_{i})$ and the highlight $(H_{i})$ of the $i{th}$ sample. Here, $w_{ij}$ are the tokenized words from $N_{i}$, and $w_{ik}$ are the tokenized words from $H_{i}$.

The T5 model has been successfully applied to various downstream tasks; therefore, we select it as our pre-trained model and fine-tune it for the evidence generation task. The news articles are fed into the encoder, and the extracted evidence is provided to the decoder for supervised fine-tuning. We use the cross-entropy loss function to update the model parameters during backpropagation. The overall architecture of our evidence-based model is illustrated in Figure~1.

\begin{figure}
\centering
\includegraphics[scale=.65]{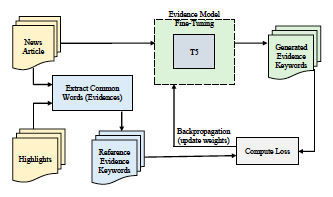}
\caption{Evidence Model - Fine-tuning T5 in CNN/DM (News articles, Highlights)}
\end{figure}

\section{Evaluation Detail}

\subsection{Intrinsic Evaluation}
To determine the similarity between our evidence-based queries and the original queries, we performed an intrinsic evaluation using the open-source library spaCy\footnote{https://spacy.io/}.

\subsection{Extrinsic Evaluation}

\subsubsection{Summarization using Pre-trained Models}
After generating the evidence-based queries, we ranked the sentences in each document according to their relevance to the generated queries, thereby transforming the documents into query-relevant inputs. This ranking ensures that query-related sentences appear at the beginning of the document, reducing the risk that important information is truncated due to input size limitations. Finally, we fine-tuned several pre-trained summarization models (four models were used separately in our experiments) on the query-relevant documents from the Debatepedia dataset to generate query-focused summaries.

{\bf Sentence Ranking}

We used the Debatepedia dataset for our QFS task, where sentence ranking helps prepare documents as query-relevant text inputs. First, we applied the evidence-based model to generate evidence-based queries for the documents in the Debatepedia dataset. Then, for each sample, we split the document into a list of sentences and converted all texts—including the generated evidence-based query and the document sentences—into their corresponding vector representations using Equations 2, 3, and 4, respectively. Next, we computed the similarity between each sentence and the query using spaCy’s similarity metric, as shown in Equation 5. Finally, we sorted all sentences in descending order of their similarity scores to construct a query-relevant document, as described in Equation~6.

\begin{equation}
S_{i} = sentenceTokenization(D_{i}) 
\end{equation}
\begin{equation}
E_{i}^{vec} = Doc2Vec(E_{i}) 
\end{equation}
\begin{equation}
s_{j}^{vec} = Doc2Vec(s_{j}) ; [s_{j} \in S_{i}]
\end{equation}
\begin{equation}
s_{j}^{sim} = spaCy.similarity(E_{i}^{vec}, s_{j}^{vec}) 
\end{equation}
\begin{equation}
D_{i}^{E} = \{s_{1}, s_{2}, \ldots, s_{p}, s_{q}, \ldots, s_{|D_{i}|}\}
\end{equation}
where $S_{i}$ is the list of sentences of the document $D_{i}$ in the $i^{th}$ sample. $E_{i}^{vec}$ and $s_{j}^{vec}$ are the vector representations of the evidence $E_{i}$ and the sentence $s_{j}$, respectively. $s_{j}^{sim}$ is the similarity score between evidence-based query and the $j^{th}$ sentence of the document. Finally, $D_{i}^{E}$ represents the query focused document of the $i^{th}$ sample where $[ \forall p, q \; s_{p}^{sim} \geq s_{q}^{sim}, p < q]$

{\bf Summarization Model}

We used transformer-based pre-trained models for our summarization task: PEGASUS, BART, RoBERTa, and LED. Since these models can handle a limited number of input tokens (1024 tokens for PEGASUS, BART, and LED, and 514 tokens for RoBERTa), it is important to place the most query-relevant tokens at the beginning of the input sequence during training to generate query-focused summaries effectively. Our sentence-ranking approach ensures that the most query-relevant sentences appear at the beginning of the input.

Pre-trained models are typically trained on specific downstream tasks and can be fine-tuned for similar tasks on different datasets. We selected PEGASUS, BART, RoBERTa, and LED because they are pre-trained for summarization or sentence generation tasks. We then fine-tuned these models on query-focused documents from the Debatepedia dataset for the QFS task to evaluate our hypotheses.

\subsubsection{Summarization using a SOTA QFS Model}
We conducted another experiment using Query-Sum \citep{xu-lapata-2020-coarse}, a recent state-of-the-art (SOTA) QFS model, to compare the results obtained using the original queries and the evidence-based queries on the TD-QFS dataset.

\section{Experimental Setup}

\subsection{Datasets}
We used the CNN/DailyMail dataset to train the evidence-based model, and the Debatepedia and TD-QFS datasets to evaluate it using both the generated evidence-based queries and the original queries.

\subsection{Implementation Details}
We used 70K training and 1,337 validation samples from the CNN/DailyMail dataset to fine-tune the evidence-based model. For the summarization task, we used 12K training and 719 validation samples from the Debatepedia dataset, along with five pre-trained models: T5, PEGASUS, BART, RoBERTa, and LED.

We used a similar parameter configuration to fine-tune both the evidence generation and summarization models. We set the number of epochs to 3, weight decay to 0.01, and learning rate to 5e-05. We used the Adam optimizer with $\beta_1 = 0.9$, $\beta_2 = 0.999$, and $\epsilon = 1e^{-08}$. The training batch size was set to 8, and the evaluation batch size to 32. During fine-tuning of the evidence-based model, we set warmup steps to 5,000 and evaluated the model every 500 steps. For the summarization models, we set warmup steps to 1,000 and evaluated the models every 250 steps.

To implement the QuerySum model, we followed the instructions provided by \citet{xu-lapata-2020-coarse} for both experimental settings, using original and evidence-based queries, respectively.

\section{Results and Discussion}
After performing the intrinsic evaluation, we computed the similarity scores between the original queries and the evidence-based queries for the Debatepedia and TD-QFS datasets, as shown in Table~2. The Debatepedia dataset yields a lower similarity score since its queries are formulated as questions. In contrast, both TD-QFS queries and our evidence-based queries are represented as sets of keywords.

\begin{table}
\centering
\begin{tabular}{|c|c|} \hline
Dataset & Similarity Score \\ \hline
Debatepedia & 0.69 \\ 
TD-QFS & 0.87 \\ \hline
\end{tabular}
\caption{Similarity score between original query and the evidence-based query}
\end{table}

We did not aim to achieve state-of-the-art results on the Debatepedia dataset for the QFS task; rather, our goal was to demonstrate that evidence-based queries can perform better than the original queries available in the dataset across four different pre-trained models. Our experimental results are shown in Table~3.

\begin{table}
\centering
\begin{tabular}{|c|c|c|c|c|} \hline
MODEL & Query & R-1 & R-2 & R-L \\ \hline
\multirow{2}{*}{Pegasus} & Original & 25.75 & 9.64 & 23.70 \\ 
                                         &  Evidence & 26.35 & 9.71 & 24.10 \\ \hline
\multirow{2}{*}{BART} & Original & 39.26 & 24.87 & 37.81 \\ 
                                    & Evidence & 40.41 & 26.03 & 39.12 \\ \hline
\multirow{2}{*}{RoBERTa} & Original & 19.35 & 2.78 & 17.81 \\ 
                                          & Evidence & 19.55 & 3.09 & 18.22 \\ \hline
\multirow{2}{*}{LED} & Original & 37.73 & 22.34 & 36.13 \\ 
                                 & Evidence & 38.08 & 22.41 & 36.48 \\ \hline
\end{tabular}
\caption{Performance of Pegasus, BART, RoBERTa and LED models on Debatepedia dataset. R-1, R-2 and R-L stand for the F1 score of ROUGE 1, 2, and L, respectively}
\end{table}

From Table~3, we observe that our evidence-based queries consistently outperform the original dataset queries in generating summaries across all four pre-trained models. BART achieves the highest ROUGE-1, ROUGE-2, and ROUGE-L scores among the four models. LED shows the second-best performance, while RoBERTa obtains the lowest scores, except for the precision values of ROUGE-1 and ROUGE-L.

By implementing the same setup as the Query-Sum model, we obtained ROUGE scores that differ from those reported in the original paper. Therefore, we report the results obtained from our own experimental environment. After replacing the original queries with evidence-based queries, we achieved a higher ROUGE-SU4 score, while ROUGE-1 and ROUGE-2 remained very close to the original scores. These results are presented in Table~4.

\begin{table}
\centering
\begin{tabular}{|c|c|c|c|} \hline
Query & R-1 & R-2 & R-SU4 \\ \hline
Original & 36.22 & 10.29 & 15.19 \\ 
Evidence & 35.95 & 9.97 & 15.94 \\ \hline
\end{tabular}
\caption{Performance using QuerySum \citep{xu-lapata-2020-coarse} model on TD-QFS dataset. R-1, R-2 and R-SU4 stand for the F1 score of ROUGE 1, 2, and SU4, respectively}
\end{table}

Based on the results in Tables 3 and 4, we conclude that our evidence-based model successfully replaces the original queries in the Debatepedia and TD-QFS datasets for the QFS task, achieving improved performance. Hence, this evidence-based model can support query-free summarization datasets by generating queries from their documents for the QFS task.

\section{Conclusion}
In this work, we present an evidence-based query generation model and provide comparative evidence that our approach successfully helps summarization models generate better summaries for query-focused datasets. In the future, we would like to extend our query generation approach to large-scale query-free datasets and further investigate how the generated queries support the QFS task.

\bibliography{anthology}

@inproceedings{lin-2004-rouge,
    title = "{ROUGE}: A Package for Automatic Evaluation of Summaries",
    author = "Lin, Chin-Yew",
    booktitle = "Text Summarization Branches Out",
    month = jul,
    year = "2004",
    address = "Barcelona, Spain",
    publisher = "Association for Computational Linguistics",
    url = "https://aclanthology.org/W04-1013",
    pages = "74--81",
}

@inproceedings{gupta-etal-2007-measuring,
    title = "Measuring Importance and Query Relevance in Topic-focused Multi-document Summarization",
    author = "Gupta, Surabhi  and
      Nenkova, Ani  and
      Jurafsky, Dan",
    booktitle = "Proceedings of the 45th Annual Meeting of the Association for Computational Linguistics Companion Volume Proceedings of the Demo and Poster Sessions",
    month = jun,
    year = "2007",
    address = "Prague, Czech Republic",
    publisher = "Association for Computational Linguistics",
    url = "https://aclanthology.org/P07-2049",
    pages = "193--196",
}

@inproceedings{10.5555/1625275.1625743, author = {Wan, Xiaojun and Yang, Jianwu and Xiao, Jianguo}, title = {Manifold-Ranking Based Topic-Focused Multi-Document Summarization}, year = {2007}, publisher = {Morgan Kaufmann Publishers Inc.}, address = {San Francisco, CA, USA}, booktitle = {Proceedings of the 20th International Joint Conference on Artifical Intelligence}, pages = {2903–2908}, numpages = {6}, location = {Hyderabad, India}, series = {IJCAI'07} }

@article{OUYANG2011227,
title = {Applying regression models to query-focused multi-document summarization},
journal = {Information Processing \& Management},
volume = {47},
number = {2},
pages = {227-237},
year = {2011},
issn = {0306-4573},
doi = {https://doi.org/10.1016/j.ipm.2010.03.005},
url = {https://www.sciencedirect.com/science/article/pii/S0306457310000257},
author = {You Ouyang and Wenjie Li and Sujian Li and Qin Lu}
}

@inproceedings{10.1145/3077136.3080690, author = {Feigenblat, Guy and Roitman, Haggai and Boni, Odellia and Konopnicki, David}, title = {Unsupervised Query-Focused Multi-Document Summarization Using the Cross Entropy Method}, year = {2017}, isbn = {9781450350228}, publisher = {Association for Computing Machinery}, address = {New York, NY, USA}, url = {https://doi.org/10.1145/3077136.3080690}, doi = {10.1145/3077136.3080690}, booktitle = {Proceedings of the 40th International ACM SIGIR Conference on Research and Development in Information Retrieval}, pages = {961–964}, numpages = {4}, location = {Shinjuku, Tokyo, Japan}, series = {SIGIR '17} }

@inproceedings{nema-etal-2017-diversity,
    title = "Diversity driven attention model for query-based abstractive summarization",
    author = "Nema, Preksha  and
      Khapra, Mitesh M.  and
      Laha, Anirban  and
      Ravindran, Balaraman",
    booktitle = "Proceedings of the 55th Annual Meeting of the Association for Computational Linguistics (Volume 1: Long Papers)",
    month = jul,
    year = "2017",
    address = "Vancouver, Canada",
    publisher = "Association for Computational Linguistics",
    url = "https://aclanthology.org/P17-1098",
    doi = "10.18653/v1/P17-1098",
    pages = "1063--1072",
}

@misc{hasselqvist2017querybased,
      title={Query-Based Abstractive Summarization Using Neural Networks}, 
      author={Johan Hasselqvist and Niklas Helmertz and Mikael Kågebäck},
      year={2017},
      eprint={1712.06100},
      archivePrefix={arXiv},
      primaryClass={cs.CL}
}

@misc{baumel2018query,
      title={Query Focused Abstractive Summarization: Incorporating Query Relevance, Multi-Document Coverage, and Summary Length Constraints into seq2seq Models}, 
      author={Tal Baumel and Matan Eyal and Michael Elhadad},
      year={2018},
      eprint={1801.07704},
      archivePrefix={arXiv},
      primaryClass={cs.CL}
}

@inproceedings{abdullah-chali-2020-towards,
    title = "Towards Generating Query to Perform Query Focused Abstractive Summarization using Pre-trained Model",
    author = "Abdullah, Deen Mohammad  and
      Chali, Yllias",
    booktitle = "Proceedings of the 13th International Conference on Natural Language Generation",
    month = dec,
    year = "2020",
    address = "Dublin, Ireland",
    publisher = "Association for Computational Linguistics",
    url = "https://aclanthology.org/2020.inlg-1.11",
    pages = "80--85",
}

@inproceedings{xu-lapata-2020-coarse,
    title = "Coarse-to-Fine Query Focused Multi-Document Summarization",
    author = "Xu, Yumo  and
      Lapata, Mirella",
    booktitle = "Proceedings of the 2020 Conference on Empirical Methods in Natural Language Processing (EMNLP)",
    month = nov,
    year = "2020",
    address = "Online",
    publisher = "Association for Computational Linguistics",
    url = "https://aclanthology.org/2020.emnlp-main.296",
    doi = "10.18653/v1/2020.emnlp-main.296",
    pages = "3632--3645",
}

@InProceedings{10.1007/978-3-030-47358-7_35,
author="Laskar, Md Tahmid Rahman
and Hoque, Enamul
and Huang, Jimmy",
editor="Goutte, Cyril
and Zhu, Xiaodan",
title="Query Focused Abstractive Summarization via Incorporating Query Relevance and Transfer Learning with Transformer Models",
booktitle="Advances in Artificial Intelligence",
year="2020",
publisher="Springer International Publishing",
address="Cham",
pages="342--348",
isbn="978-3-030-47358-7"
}

@misc{su2021improve,
      title={Improve Query Focused Abstractive Summarization by Incorporating Answer Relevance}, 
      author={Dan Su and Tiezheng Yu and Pascale Fung},
      year={2021},
      eprint={2105.12969},
      archivePrefix={arXiv},
      primaryClass={cs.CL}
}

@inproceedings{fisher2006query,
  title={Query-focused summarization by supervised sentence ranking and skewed word distributions},
  author={Fisher, Seeger and Roark, Brian},
  booktitle={Proceedings of the Document Understanding Conference, DUC-2006, New York, USA},
  year={2006}
}

@misc{see2017point,
      title={Get To The Point: Summarization with Pointer-Generator Networks}, 
      author={Abigail See and Peter J. Liu and Christopher D. Manning},
      year={2017},
      eprint={1704.04368},
      archivePrefix={arXiv},
      primaryClass={cs.CL}
}

@inproceedings{liu-lapata-2019-text,
    title = "Text Summarization with Pretrained Encoders",
    author = "Liu, Yang  and
      Lapata, Mirella",
    booktitle = "Proceedings of the 2019 Conference on Empirical Methods in Natural Language Processing and the 9th International Joint Conference on Natural Language Processing (EMNLP-IJCNLP)",
    month = nov,
    year = "2019",
    address = "Hong Kong, China",
    publisher = "Association for Computational Linguistics",
    url = "https://aclanthology.org/D19-1387",
    doi = "10.18653/v1/D19-1387",
    pages = "3730--3740",
}

@inproceedings{devlin-etal-2019-bert,
    title = "{BERT}: Pre-training of Deep Bidirectional Transformers for Language Understanding",
    author = "Devlin, Jacob  and
      Chang, Ming-Wei  and
      Lee, Kenton  and
      Toutanova, Kristina",
    booktitle = "Proceedings of the 2019 Conference of the North {A}merican Chapter of the Association for Computational Linguistics: Human Language Technologies, Volume 1 (Long and Short Papers)",
    month = jun,
    year = "2019",
    address = "Minneapolis, Minnesota",
    publisher = "Association for Computational Linguistics",
    url = "https://aclanthology.org/N19-1423",
    doi = "10.18653/v1/N19-1423",
    pages = "4171--4186",
}

@article{radford2019language,
  title={Language models are unsupervised multitask learners},
  author={Radford, Alec and Wu, Jeffrey and Child, Rewon and Luan, David and Amodei, Dario and Sutskever, Ilya and others},
  journal={OpenAI blog},
  volume={1},
  number={8},
  pages={9},
  year={2019}
}

@misc{liu2019roberta,
      title={RoBERTa: A Robustly Optimized BERT Pretraining Approach}, 
      author={Yinhan Liu and Myle Ott and Naman Goyal and Jingfei Du and Mandar Joshi and Danqi Chen and Omer Levy and Mike Lewis and Luke Zettlemoyer and Veselin Stoyanov},
      year={2019},
      eprint={1907.11692},
      archivePrefix={arXiv},
      primaryClass={cs.CL}
}

@misc{raffel2020exploring,
      title={Exploring the Limits of Transfer Learning with a Unified Text-to-Text Transformer}, 
      author={Colin Raffel and Noam Shazeer and Adam Roberts and Katherine Lee and Sharan Narang and Michael Matena and Yanqi Zhou and Wei Li and Peter J. Liu},
      year={2020},
      eprint={1910.10683},
      archivePrefix={arXiv},
      primaryClass={cs.LG}
}

@misc{beltagy2020longformer,
      title={Longformer: The Long-Document Transformer}, 
      author={Iz Beltagy and Matthew E. Peters and Arman Cohan},
      year={2020},
      eprint={2004.05150},
      archivePrefix={arXiv},
      primaryClass={cs.CL}
}

@inproceedings{lewis-etal-2020-bart,
    title = "{BART}: Denoising Sequence-to-Sequence Pre-training for Natural Language Generation, Translation, and Comprehension",
    author = "Lewis, Mike  and
      Liu, Yinhan  and
      Goyal, Naman  and
      Ghazvininejad, Marjan  and
      Mohamed, Abdelrahman  and
      Levy, Omer  and
      Stoyanov, Veselin  and
      Zettlemoyer, Luke",
    booktitle = "Proceedings of the 58th Annual Meeting of the Association for Computational Linguistics",
    month = jul,
    year = "2020",
    address = "Online",
    publisher = "Association for Computational Linguistics",
    url = "https://aclanthology.org/2020.acl-main.703",
    doi = "10.18653/v1/2020.acl-main.703",
    pages = "7871--7880",
}

@InProceedings{pmlr-v119-zhang20ae,
  title = 	 {{PEGASUS}: Pre-training with Extracted Gap-sentences for Abstractive Summarization},
  author =       {Zhang, Jingqing and Zhao, Yao and Saleh, Mohammad and Liu, Peter},
  booktitle = 	 {Proceedings of the 37th International Conference on Machine Learning},
  pages = 	 {11328--11339},
  year = 	 {2020},
  editor = 	 {III, Hal Daumé and Singh, Aarti},
  volume = 	 {119},
  series = 	 {Proceedings of Machine Learning Research},
  month = 	 {13--18 Jul},
  publisher =    {PMLR},
  url = 	 {https://proceedings.mlr.press/v119/zhang20ae.html}
}

@inproceedings{ma-huang-2006-uniform,
    title = "Uniform and Effective Tagging of a Heterogeneous Giga-word Corpus",
    author = "Ma, Wei-Yun  and
      Huang, Chu-Ren",
    booktitle = "Proceedings of the Fifth International Conference on Language Resources and Evaluation ({LREC}{'}06)",
    month = may,
    year = "2006",
    address = "Genoa, Italy",
    publisher = "European Language Resources Association (ELRA)",
    url = "http://www.lrec-conf.org/proceedings/lrec2006/pdf/294_pdf.pdf",
}

@article{hermann2015teaching,
  title={Teaching machines to read and comprehend},
  author={Hermann, Karl Moritz and Kocisky, Tomas and Grefenstette, Edward and Espeholt, Lasse and Kay, Will and Suleyman, Mustafa and Blunsom, Phil},
  journal={Advances in neural information processing systems},
  volume={28},
  year={2015}
}

@inproceedings{rajpurkar-etal-2016-squad,
    title = "{SQ}u{AD}: 100,000+ Questions for Machine Comprehension of Text",
    author = "Rajpurkar, Pranav  and
      Zhang, Jian  and
      Lopyrev, Konstantin  and
      Liang, Percy",
    booktitle = "Proceedings of the 2016 Conference on Empirical Methods in Natural Language Processing",
    month = nov,
    year = "2016",
    address = "Austin, Texas",
    publisher = "Association for Computational Linguistics",
    url = "https://aclanthology.org/D16-1264",
    doi = "10.18653/v1/D16-1264",
    pages = "2383--2392",
}

@article{Baumel_Cohen_Elhadad_2016, title={Topic Concentration in Query Focused Summarization Datasets}, volume={30}, url={https://ojs.aaai.org/index.php/AAAI/article/view/10323}, DOI={10.1609/aaai.v30i1.10323}, abstractNote={ &lt;p&gt; Query-Focused Summarization (QFS) summarizes a document cluster in response to a specific input query. QFS algorithms must combine query relevance assessment, central content identification, and redundancy avoidance. Frustratingly, state of the art algorithms designed for QFS do not significantly improve upon generic summarization methods, which ignore query relevance, when evaluated on traditional QFS datasets. We hypothesize this lack of success stems from the nature of the dataset. We define a task-based method to quantify topic concentration in datasets, i.e., the ratio of sentences within the dataset that are relevant to the query, and observe that the DUC 2005, 2006 and 2007 datasets suffer from very high topic concentration. We introduce TD-QFS, a new QFS dataset with controlled levels of topic concentration. We compare competitive baseline algorithms on TD-QFS and report strong improvement in ROUGE performance for algorithms that properly model query relevance as opposed to generic summarizers. We further present three new and simple QFS algorithms, RelSum, ThresholdSum, and TFIDF-KLSum that outperform state of the art QFS algorithms on the TD-QFS dataset by a large margin. &lt;/p&gt; }, number={1}, journal={Proceedings of the AAAI Conference on Artificial Intelligence}, author={Baumel, Tal and Cohen, Raphael and Elhadad, Michael}, year={2016}, month={Mar.} }

@inproceedings{rush-etal-2015-neural,
    title = "A Neural Attention Model for Abstractive Sentence Summarization",
    author = "Rush, Alexander M.  and
      Chopra, Sumit  and
      Weston, Jason",
    booktitle = "Proceedings of the 2015 Conference on Empirical Methods in Natural Language Processing",
    month = sep,
    year = "2015",
    address = "Lisbon, Portugal",
    publisher = "Association for Computational Linguistics",
    url = "https://aclanthology.org/D15-1044",
    doi = "10.18653/v1/D15-1044",
    pages = "379--389",
}

@inproceedings{nallapati-etal-2016-abstractive,
    title = "Abstractive Text Summarization using Sequence-to-sequence {RNN}s and Beyond",
    author = "Nallapati, Ramesh  and
      Zhou, Bowen  and
      dos Santos, Cicero  and
      G?l{\c{c}}ehre, {\c{C}}a{\u{g}}lar  and
      Xiang, Bing",
    booktitle = "Proceedings of the 20th {SIGNLL} Conference on Computational Natural Language Learning",
    month = aug,
    year = "2016",
    address = "Berlin, Germany",
    publisher = "Association for Computational Linguistics",
    url = "https://aclanthology.org/K16-1028",
    doi = "10.18653/v1/K16-1028",
    pages = "280--290",
}

@inproceedings{durrett-etal-2016-learning,
    title = "Learning-Based Single-Document Summarization with Compression and Anaphoricity Constraints",
    author = "Durrett, Greg  and
      Berg-Kirkpatrick, Taylor  and
      Klein, Dan",
    booktitle = "Proceedings of the 54th Annual Meeting of the Association for Computational Linguistics (Volume 1: Long Papers)",
    month = aug,
    year = "2016",
    address = "Berlin, Germany",
    publisher = "Association for Computational Linguistics",
    url = "https://aclanthology.org/P16-1188",
    doi = "10.18653/v1/P16-1188",
    pages = "1998--2008",
}

\appendix

\onecolumn

\section{Appendix}
In Table~5, we are showing the generated queries and the original queries of the TD-QFS data set, where sample number 0 2 denotes the sample belongs to document number 2 of cluster number 0. 
\begin{table}[H]
\centering
\begin{tabular}{cll}
Sample Number  & Original Query  & Evidence-based Query  \\
0 0  & asthma causes  & asthma chronic disease affects air   \\
0 1  & asthma treatment  & asthma chronic lung disease inflam  \\
0 2  & exercise induced asthma  & exercise induced asthma relatively common  \\
0 3  & atopic dermatitis  & atopic dermatiti  \\
0 4  & atopic dermatitis  & atopic dermatiti  \\
0 5  & asthma allergy  & asthma chronic lung disease inflam  \\
0 6  & asthma medication  & right medications asthma depend number things  \\
0 7  & exercise for asthmatic  & exercise help prevent asthma attacks control  \\
0 8  & atopic dermatitis medications  & atopic dermatiti  \\
1 0  & salt obesity  & salt intake cause obesity people adults  \\
1 1  & obesity screening  & body mass index measure used determine  \\
1 2  & childhood obesity  & childhood obesity serious medical condition affect  \\
1 3  & causes of childhood obesity  & children worldwide either overweight obese there  \\
1 4  & obesity and lifestyle change  & obesity become increasing epidemic affects  \\
1 5  & obesity and diabetes  & almost people type diabetes overweight number  \\
1 6  & retaining fluid  water & retention also known medical term  \\
1 7  & body mass index  & body mass index simple mathematical formula  \\
1 8  & childhood obesity low income  & childhood obesity rates increased percent  \\
1 9  & obesity and nutrition  & americans eating foods high calories  \\
1 10  & obesity metabolism  & researchers mice born without section gene  \\
1 11  & emergence of type 2 diabetes  & type diabetes caused combination factors including  \\
2 0  & Lung cancer  & there basically types lung cancer there  \\
2 1  & Lung cancer diagnosis  & lung cancer screening currently routine practice  \\
2 2  & Stage of lung cancer  & stage refers extent lung cancer  \\
2 3  & Types of lung cancer  & sclc accounts  \\
2 4  & Lung Cancer in Women  & smoking contributes lung cancer cigarettes  \\
2 5  & Lung Cancer chemotherapy  & chemotherapy main treatment small cell lung  \\
2 6  & Lung cancer causes  & lung cancer uncontrolled cell growth  \\
2 7  & Lung cancer treatment  & lung cancer forms tissues usually cells \\ 
2 8  & Lung cancer staging tests  & staging helps doctor decide kind treatment  \\
2 9  & Non-small cell lung cancer & treatment  non-small cell lung cancer  \\
2 10  & Risk factors for Lung Cancer in Women  & women account lung cancer cases women  \\
3 0  & alzheimer memory  & alzheimer common form  \\
3 1  & cognitive impairment  & mild cognitive impairment mild cognitive impairment  \\
3 2  & alzheimer's symptoms  & alzheimer symptoms include  \\
3 3  & semantic dementia  & semantic dementia progressive neurodegenerative  \\
3 4  & helping retrieve memory alzheimer & experimental drug could help recover memory  \\
3 5  & Vascular Dementia  & vascular dementia affects different  \\
3 6  & alzheimer diagnosis  & alzheimer considered disease  \\
3 7  & first symptoms dementia  & dementia specific disease describes wide range  \\
\end{tabular}
\caption{Original and the evidence-based queries from TD-QFS dataset, where sample number 0 2 indicates that the sample belongs to document number 2 in cluster number 0.}
\end{table}

\end{document}